\documentclass[11pt]{article}
\usepackage{coling2020}
\usepackage{times}
\usepackage{url}
\usepackage{latexsym}
\usepackage{tabulary}
\usepackage{booktabs}       
\usepackage{graphicx}
\usepackage{pgfplots}
\pgfplotsset{width=10cm,compat=1.9}
\usepackage{caption}
\usepackage{amsmath}
\usepackage{amsfonts}       

\colingfinalcopy 

\title{Better Sign Language Translation with STMC-Transformer}

\author{Kayo Yin\thanks{*Work carried out while at École Polytechnique.} \\
  Language Technologies Institute\\
  Carnegie Mellon University \\
  {\tt kayo@cmu.edu} \\\And
  Jesse Read \\
  LIX, Ecole Polytechnique \\
  Institut Polytechnique de Paris \\
  {\tt jesse.read@polytechnique.edu} \\}

\date{}

\begin{document}
\maketitle
\begin{abstract}
Sign Language Translation (SLT) first uses a Sign Language Recognition (SLR) system to extract sign language glosses from videos. Then, a translation system generates spoken language translations from the sign language glosses. This paper focuses on the translation system and introduces the STMC-Transformer which improves on the current state-of-the-art by over 5 and 7 BLEU respectively on gloss-to-text and video-to-text translation of the PHOENIX-Weather 2014T dataset. On the ASLG-PC12 corpus, we report an increase of over 16 BLEU. 

We also demonstrate the problem in current methods that rely on gloss supervision. The video-to-text translation of our STMC-Transformer outperforms translation of GT glosses. This contradicts previous claims that GT gloss translation acts as an upper bound for SLT performance and reveals that glosses are an inefficient representation of sign language. For future SLT research, we therefore suggest an end-to-end training of the recognition and translation models, or using a different sign language annotation scheme.
\end{abstract}

\section{Introduction}
\label{intro}

%
%
\blfootnote{
    %
    %
    %
    %
    %
    %
     \hspace{-0.65cm}  
     This work is licensed under a Creative Commons 
     Attribution 4.0 International License.
     License details:
     \url{http://creativecommons.org/licenses/by/4.0/}.
}

Communication holds a central position in our daily lives and social interactions. Yet, in a predominantly aural society, sign language users are often deprived of effective communication. Deaf people face daily issues of social isolation and miscommunication to this day \cite{SOUZA2017}. This paper is motivated to provide assistive technology that allow Deaf people to communicate in their own language.

In general, sign languages developed independently of spoken language and do not share the grammar of their spoken counterparts \cite{stokoe}. For this, Sign Language Recognition (SLR) systems on their own cannot capture the underlying grammar and complexities of sign language, and Sign Language Translation (SLT) faces the additional challenge of taking into account the unique linguistic features during translation.

\begin{center}
\center
    \includegraphics[width=.8\linewidth]{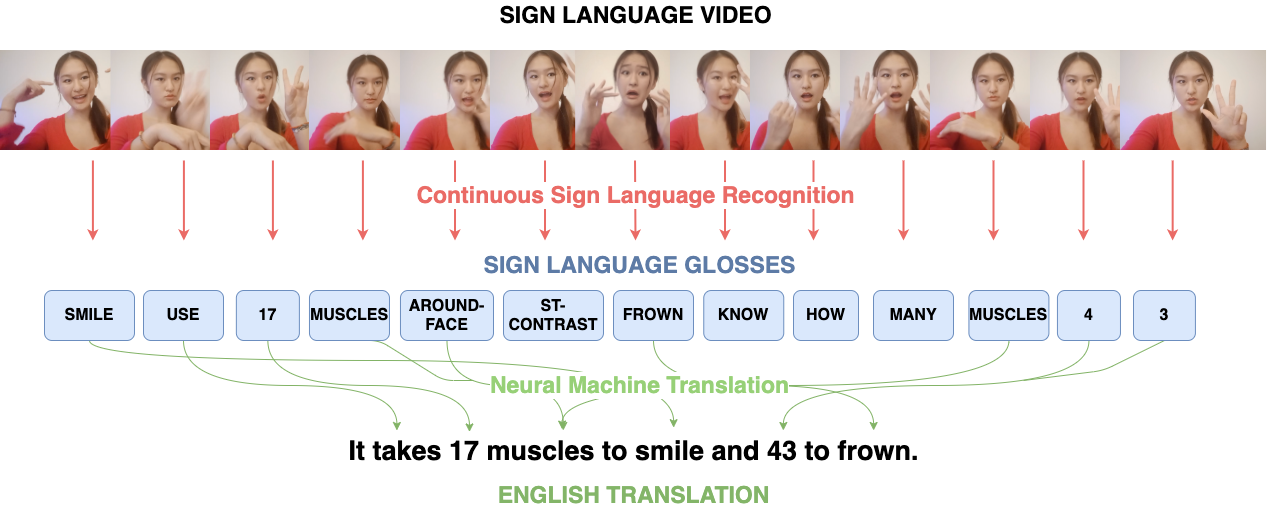}
    \captionof{figure}{Sign language translation pipeline\footnotemark.}
    \label{fig:slt}
\end{center}  
\footnotetext{Gloss annotation from \url{https://www.handspeak.com/translate/index.php?id=288}}

As shown in Figure \ref{fig:slt}, current SLT approaches involve two steps. First, a tokenization system generates glosses from sign language videos. Then, a translation system translates the recognized glosses into spoken language. Recent work \cite{Orbay2020NeuralSL,stmc} has addressed the first step, but there has been none improving the translation system. This paper aims to fill this research gap by leveraging recent success in Neural Machine Translation (NMT), namely Transformers. 

Another limit to current SLT models is that they use glosses as an intermediate representation of sign language. We show that having a perfect continuous SLR system will not necessarily improve SLT results. We introduce the STMC-Transformer model performing video-to-text translation that surpasses translation of ground truth glosses, which reveals that glosses are a flawed representation of sign language.

The contributions of this paper can be summarized as:
\begin{enumerate}
  \item A novel STMC-Transformer model for video-to-text translation surpassing GT glosses translation contrary to previous assumptions
  \item The first successful application of Transformers to SLT achieving state-of-the-art results in both gloss to text and video to text translation on PHOENIX-Weather 2014T and ASLG-PC12 datasets
  \item The first usage of weight tying, transfer learning, and ensemble learning in SLT and a comprehensive series of baseline results with Transformers to underpin future research

\end{enumerate}

\section{Methods}
Despite considerable advancements made in machine translation (MT) between spoken languages, sign language processing falls behind for many reasons. Unlike spoken language, sign language is a multidimensional form of communication that relies on both manual and non-manual cues which presents additional computer vision challenges \cite{nonmanual}. These cues may occur simultaneously whereas spoken language follows a linear pattern where words are processed one at a time. Signs also vary in both space and time and the number of video frames associated to a single sign is not fixed either.

\subsection{Sign Language Glossing}
Glossing corresponds to transcribing sign language word-for-word by means of another written language. Glosses differ from translation as they merely indicate what each part in a sign language sentence mean, but do not form an appropriate sentence in the spoken language. While various sign language corpus projects have provided different guidelines for gloss annotation \cite{sign1,sign2}, there is no universal standard which hinders the easy exchange of data between projects and consistency between different sign language corpora. Gloss annotations are also an imprecise representation of sign language and can lead to an information bottleneck when representing the multi-channel sign language by a single-dimensional stream of glosses.

\subsection{Sign Language Recognition}

SLR consists of identifying isolated single signs from videos. Continuous sign language recognition (CSLR) is a relatively more challenging task that identifies a sequence of running glosses from a running video. Works
in SLR and CSLR, however, only perform visual recognition and ignore the underlying linguistic features of sign language.

\subsection{Sign Language Translation}
As illustrated in Figure \ref{fig:slt}, the SLT system takes CSLR as a first step to tokenize the input video into glosses. Then, an additional step translates the glosses into a valid sentence in the target language. SLT is novel and difficult compared to other translation problems because it involves two steps: extract meaningful features from a video of a multi-cue language accurately then generate translations from an intermediate gloss representation, instead of translation from the source language directly.

\section{Related Work}
\label{sec:headings}

\subsection{Sign Language Recognition}

Early approaches for SLR rely on hand-crafted features \cite{sift,chinese} and use Hidden Markov Models \cite{hmm1} or Dynamic Time Warping \cite{dtw} to model sequential dependencies. More recently, 2D convolutional neural networks (2D-CNN) and 3D convolutional neural networks (3D-CNN)   effectively model spatio-temporal representations from sign language videos \cite{rcnn,rcnn3d}.

Most existing work on CSLR divides the task into three sub-tasks: alignment learning, single-gloss SLR, and sequence construction \cite{resign,hmmdtw} while others perform the task in an end-to-end fashion using deep learning \cite{3Dcnn,subunet}.

\subsection{Sign Language Translation}
SLT was formalized in \newcite{camgozslt} where they introduce the PHOENIX-Weather 2014T dataset and jointly use a 2D-CNN model to extract gloss-level features from video frames, and a seq2seq model to perform German sign language translation. Subsequent works on this dataset \cite{Orbay2020NeuralSL,stmc} all focus on improving the CSLR component in SLT. A contemporaneous paper \cite{camgoz2020sign} also obtains encouraging results with multi-task Transformers for both tokenization and translation, however their CSLR performance is sub-optimal, with a higher Word Error Rate than baseline models.

Similar work has been done on Korean sign language by \newcite{keti} where they estimate human keypoints to extract glosses, then use seq2seq models for translation. \newcite{aslseq} use seq2seq models to translate ASL glosses of the ASLG-PC12 dataset \cite{asl}.

\subsection{Neural Machine Translation}
Neural Machine Translation (NMT) employs neural networks to carry out automated text translation. Recent methods typically use an encoder-decoder architecture, also known as seq2seq models.

Earlier approaches use recurrent \cite{rnn1,rnn2} and convolutional networks \cite{cnn1,cnn2} for the encoder and the decoder. However, standard seq2seq networks are unable to model long-term dependencies in large input sentences without causing an information bottleneck. To address this issue, recent works use attention mechanisms \cite{bahdanau,luong} that calculates context-dependent alignment scores between encoder and decoder hidden states. \newcite{transformer} introduces the Transformer, a seq2seq model relying on self-attention that obtains state-of-the-art results in NMT.

\begin{figure*}
\center
    \includegraphics[width=\linewidth]{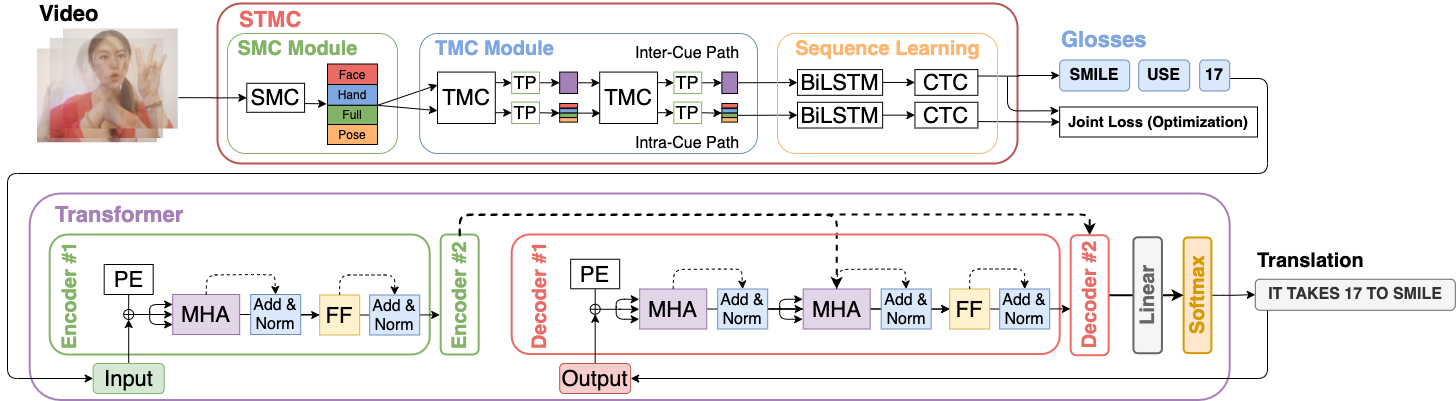}
    \caption{STMC-Transformer network for SLT. PE: Positional Encoding, MHA: Multihead Attention, FF: Feed Forward.}
    \label{fig:stmc}
\end{figure*} 

\section{Model architecture}

For translation from videos to text, we propose the STMC-Transformer network illustrated in Figure \ref{fig:stmc}.

\subsection{Spatial-Temporal Multi-Cue (STMC) Network}
Our work is the first to use STMC networks \cite{stmc} for SLT. A spatial multi-cue (SMC) module with a self-contained pose estimation branch decomposes the input video into spatial features of multiple visual cues (face, hand, full-frame and pose). Then, a temporal multi-cue (TMC) module with stacked TMC blocks and temporal pooling (TP) layers calculates temporal correlations within (inter-cue) and between cues (intra-cue) at different time steps, which preserves each unique cue while exploring their relation at the same time. The inter-cue and intra-cue features are each analyzed by Bi-directional Long Short-Term Memory (BiLSTM) \cite{rnn2} and Connectionist Temporal Classification (CTC) \cite{ctc} units for sequence learning and inference. 

This architecture efficiently processes multiple visual cues from sign language video in collaboration with each other, and achieves state-of-the-art performance on three SLR benchmarks. On the PHOENIX-Weather 2014T dataset, it achieves a Word Error Rate of 21.0 for the SLR task. 

\subsection{Transformer}

For translation, we train a two-layered Transformer to maximize the log-likelihood $$\sum_{(x_i, y_i) \in D} \log P(y_i | x_i, \theta)$$
where $D$ contains gloss-text pairs $(x_i, y_i)$. 

Two layers, compared to six in most spoken language translation, is empirically shown to be optimal in Section \ref{sec:layers}, likely because our datasets are limited in size. We refer to the original Transformer paper \cite{transformer} for more architecture details.

\section{Datasets}

\begin{center}
\resizebox{\textwidth}{!}{\begin{tabular}{lllllllllllllll}
\toprule
& \multicolumn{3}{c}{German Sign Gloss} & \multicolumn{3}{c}{German} & \multicolumn{3}{c}{American Sign Gloss} & \multicolumn{3}{c}{English} \\
\cmidrule(l){2-4}\cmidrule(l){5-7}\cmidrule(l){8-10}\cmidrule(l){11-13}
 & {Train} & {Dev}  & {Test} & {Train} & {Dev}  & {Test}& {Train} & {Dev}  & {Test}& {Train} & {Dev}  & {Test}\\
\midrule
Phrases & 7,096 & 519 & 642 & 7,096 & 519 & 642 & 82,709 & 4,000 & 1,000 & 82,709 & 4,000 & 1,000\\
Vocab. & 1,066 & 393 & 411 & 2,887 & 951 & 1,001 & 15782 & 4,323 & 2,150 & 21,600 & 5,634 & 2,609 \\
tot. words & 67,781 & 3,745 & 4,257 & 99,081 & 6,820 & 7,816 & 862,046 & 41,030 & 10,503  & 975,942 & 46,637 & 11,953\\
tot. OOVs & \textendash & 19 & 22 & \textendash & 57 & 60 & \textendash & 255 & 83 & \textendash & 369 & 99 \\
singletons & 337 & \textendash & \textendash & 1,077 & \textendash & \textendash & 6,133 & \textendash & \textendash  & 8,542 & \textendash & \textendash \\
\bottomrule
\end{tabular}} 
 \captionof{table}{Statistics of the RWTH-PHOENIX-Weather 2014T and ASLG-PC12 datasets. Out-of-vocabulary (OOV) words are those that appear in the development and testing sets, but not in the training set. Singletons are words that appear only once during training.} 
\label{table:datasets}
\end{center}

\subsection*{PHOENIX-Weather 2014T \cite{camgozslt}}  This dataset is extracted from weather forecast airings of the German tv station PHOENIX. This dataset consists of a parallel corpus of German sign language videos from 9 different signers, gloss-level annotations with a vocabulary of 1,066 different signs and translations into German spoken language with a vocabulary of 2,887 different words. It contains 7,096 training pairs, 519 development and 642 test pairs.

\subsection*{ASLG-PC12 \cite{asl}}  This dataset is constructed from English data of Project Gutenberg that has been transformed into ASL glosses following a rule-based approach. This corpus with 87,709 training pairs allows us to evaluate Transformers on a larger dataset, where deep learning models usually require lots of data. It also allows us to compare performance across different sign languages. However, the data is limited since it does not contain sign language videos, and is less complex due to being created semi-automatically. We make our data and code publicly available\footnote{\url{https://github.com/kayoyin/transformer-slt}}.

\section{Experiments and Discussions}
Our models are built using PyTorch \cite{pytorch} and Open-NMT \cite{opennmt}. We configure Transformers with word embedding size 512, gloss level tokenization, sinusoidal positional encoding, 2,048 hidden units and 8 heads. For optimization, we use Adam \cite{adam} with $\beta_1 = 0.9, \beta_2 = 0.998$, Noam learning rate schedule, 0.1 dropout, and 0.1 label smoothing. 

We evaluate on the dev set each half-epoch and employ early stopping with patience 5. During decoding, generated $\langle unk \rangle$ tokens are replaced by the source token having the highest attention weight. This is useful when $\langle unk \rangle$ symbols correspond to proper nouns that can be directly transposed between languages \cite{opennmt}. We perform a series of experiments to find the optimal setup for this novel application. We equally experiment with various techniques often used in classic NMT to SLT such as transfer learning, weight tying and ensembling to improve model performance.

For evaluation we use BLEU \cite{bleu}, ROUGE \cite{rouge} and METEOR \cite{meteor}. For BLEU, we report BLEU-1,2,3,4 scores and as ROUGE score we report the ROUGE-L F1 score. These metrics allow us to directly compare directly with previous works. METEOR is calculated in addition as it demonstrates higher correlation with human evaluation than BLEU on several MT tasks. All reported results unless otherwise specified are averaged over 10 runs with different random seeds.

We organize our experiments into two groups:
\begin{enumerate}
    \item Gloss2Text (G2T) in which we translate GT gloss annotations to simulate perfect tokenization on both PHOENIX-Weather 2014T and ASLG-PC12 
    \item Sign2Gloss2Text (S2G2T) where we perform video-to-text translation on PHOENIX-Weather 2014T with the STMC-Transformer

\end{enumerate}

\subsection{Gloss2Text (G2T)}
\label{sec:layers}
 G2T is a text-to-text translation task that is novel and challenging compared to classic translation tasks between spoken languages because of the high linguistic variance between source and target sentences, scarcity of resources, and information loss or imprecision in the source sentence itself.
 
For ASLG-PC12, many ASL glosses are English words with an added prefix so during data pre-processing we remove all such prefixes. We also set all glosses that appear less than 5 times during training as $\langle unk \rangle$ to reduce vocabulary size. 

\begin{center}
 \resizebox{0.8\textwidth}{!}{\begin{tabular}{ccccccccccccc}
\toprule
& \multicolumn{6}{c}{Raw data} & \multicolumn{6}{c}{Preprocessed data} \\ 
\cmidrule(l){2-7}\cmidrule(l){8-13}
 & \multicolumn{2}{c}{Train} & \multicolumn{2}{c}{Dev}  & \multicolumn{2}{c}{Test} & \multicolumn{2}{c}{Train} & \multicolumn{2}{c}{Dev}  & \multicolumn{2}{c}{Test} \\
 & {ASL} & {en} & {ASL} & {en} &{ASL} & {en} &{ASL} & {en} &{ASL} & {en} &{ASL} & {en}  \\
\midrule
Vocab. & 15,782 & 21,600 & 4,323 & 5,634 & 2,150 & 2,609 & 5,906 & 7,712 & 1,163 & 1,254 &  394 & 379\\
Shared vocab. & \multicolumn{2}{c}{10,048}  & \multicolumn{2}{c}{2,652} & \multicolumn{2}{c}{1,296} & \multicolumn{2}{c}{4,941} & \multicolumn{2}{c}{899} & \multicolumn{2}{c}{287} \\
BLEU-4 & \multicolumn{2}{c}{20.97} & \multicolumn{2}{c}{21.16} & \multicolumn{2}{c}{20.63} & \multicolumn{2}{c}{38.87} & \multicolumn{2}{c}{38.74} & \multicolumn{2}{c}{38.37} \\
\bottomrule
\end{tabular}} 
\captionof{table}{Statistics of the ASLG-PC12 dataset before and after preprocessing.} 
\label{table:asl}
\end{center}

Table \ref{table:asl} shows that the source and target corpora in ASLG-PC12 are more similar to each other with many shared vocabulary and a relatively high BLEU-4 score on raw data. This allows us to compare Transformer performance on a larger and less challenging dataset.

\subsubsection*{Model size}
The original Transformer in \cite{transformer} uses 6 layers for the encoder and decoder for NMT. However, our task differs from a standard MT task between two spoken languages so we first train Transformers with 1, 2, 4 and 6 encoder-decoder layers. Networks are trained with batch size 2,048 and initial learning rate 1. 

\begin{center}
\resizebox{\linewidth}{!}{\begin{tabular}{lcccccccccccc}
\toprule
& \multicolumn{6}{c}{Dev Set} & \multicolumn{6}{c}{Test Set} \\
\cmidrule(l){2-7}\cmidrule(l){8-13}
 {Layers} & {BLEU-1} & {BLEU-2}  & {BLEU-3} & {BLEU-4} & {ROUGE-L}  & {METEOR}& {BLEU-1} & {BLEU-2}  & {BLEU-3} & {BLEU-4} & {ROUGE-L}  & {METEOR}\\
 \midrule
 1 & 43.39 & 32.47 & 24.27 & 20.26 & 44.66 & 42.64 & 43.26 & 32.23 & 25.59 & 21.31 & 45.28 & 42.56 \\
 2 & \textbf{45.31} & 33.65 & \textbf{26.73} & \textbf{22.23} & 45.74 & \textbf{43.92} & \textbf{44.57} & \textbf{33.08} &  \textbf{26.14} & \textbf{21.65} & 40.47 & \textbf{42.97} \\
 4 & 44.32 & \textbf{32.87}  & 26.15 & 21.78 & \textbf{45.86} & 43.31 & 44.10 & 32.82 & 25.99 & 21.57 & \textbf{45.44} & 42.92 \\
 6 & 44.04 & 32.46 & 25.67 & 21.34 & 44.09 & 42.32 & 43.74 & 32.44 & 25.67 & 21.32 & 41.69 & 42.58\\
\bottomrule
\end{tabular}} 
 \captionof{table}{G2T performance comparison of Transformers on PHOENIX-Weather 2014T with different number of enc-dec layers.} 
\label{table:layers}
\end{center}

To choose the best model, we mainly take into account BLEU-4 as it is currently the most widely used metric in MT. We do find that our final model outperforms the other models across all metrics. Table \ref{table:layers} shows that on PHOENIX-Weather 2014T, using 2 layers obtains the highest BLEU-4. Because our dataset is much smaller than spoken language datasets, larger networks may be disadvantaged. Moreover, a smaller model has the advantage of taking up less memory and computation time. Repeating the same experiment on ASLG-PC12, we also find 2 layers to be the optimal model size. ASLG-PC12 is larger but less complex which may also explain why smaller networks are more suitable. We carry out the rest of our experiments using 2 enc-dec layers.

\subsubsection*{Embedding schemes}

\newcite{weighttie} shows that tying the input and output embeddings while training language models may provide better performance. Our decoder is in fact a language model conditioned on the encoding of the source sentence and previous outputs, we can tie the decoder embeddings by using a shared weight matrix for the input and output word embeddings.  

In addition, models are often initialized with pre-trained embeddings for transfer learning. These embeddings are typically trained in an unsupervised manner on a large corpus of text in the desired language. We perform experiments on PHOENIX-Weather 2014T using two popular word embeddings: GloVe\footnote{\url{https://deepset.ai/german-word-embeddings}} \cite{glove}, and fastText \cite{fasttext}. To the best of our knowledge, weight-tying or pre-trained embeddings have never been employed in SLT.

\begin{center}
\resizebox{.5\linewidth}{!}{\begin{tabular}{lllll}
\toprule
& {GloVe (de)} & {fastText (de)} & {GloVe (en)} & {fastText (en)}\\
 \midrule
 {Dimension} & 300 & 300 & 300 & 300 \\
 {Source match} & 0.08\% & 0.08\% & 96.23\% & 94.64\%\\
 {Target match} & 90.53\% & 94.57\% & 97.71\% & 96.32\% \\

\bottomrule

\end{tabular}} 
 \captionof{table}{German and English pre-trained embeddings statistics} 
\label{table:emb}
\end{center}

Table \ref{table:emb} shows there is only one matching token between German glosses and the pre-trained embeddings, while over 90\% of the words in the German text appear in both pre-trained embeddings. We therefore initialize pre-trained embeddings on the decoder only, and keep random initialization for the encoder. The embedding layers are fine-tuned during training.

\begin{center}
\resizebox{\linewidth}{!}{\begin{tabular}{lcccccccccccc}
\toprule
& \multicolumn{6}{c}{Dev Set} & \multicolumn{6}{c}{Test Set} \\
\cmidrule(l){2-7}\cmidrule(l){8-13}
 {WE} & {BLEU-1} & {BLEU-2}  & {BLEU-3} & {BLEU-4} & {ROUGE-L}  & {METEOR}& {BLEU-1} & {BLEU-2}  & {BLEU-3} & {BLEU-4} & {ROUGE-L}  & {METEOR}\\
 \midrule
Vanilla embedding & 45.81 & 34.06 & \textbf{27.05} & \textbf{22.49} & \textbf{46.68} & \textbf{44.35} & \textbf{45.29} & \textbf{33.74} & \textbf{26.70} & \textbf{22.22} & \textbf{46.08} & 43.75 \\
Tied decoder & \bfseries 45.90 & \bfseries 34.10 & 26.98 & 22.31 & 46.76 & 44.51 & 45.05 & 33.38 & 26.31 & 21.74 & 45.83 & 43.45\\
GloVe & 44.37 & 32.65 & 26.00 & 21.41 & 45.03 & 42.38 & 44.69 & 32.93 & 25.73 & 21.04 & 42.70 & \textbf{44.61} \\
fastText & 44.91 & 33.23 & 26.60 & 22.04 & 46.17 & 43.70 & 44.21 & 32.90 & 25.94 & 21.64 & 45.55 & 42.95\\
\bottomrule
\end{tabular}} 
\label{table:deemb}
 \captionof{table}{G2T performance comparison using different embedding schemes on PHOENIX-Weather 2014T.} 
\end{center}

Table 5 shows that the new embedding schemes do not improve performance on PHOENIX-Weather 2014T. It may be because pre-trained embeddings are shown to be more effective when used on the encoding layer \cite{pretrain}. Another possible reason is the difference between the domain of our dataset and of the corpus the embeddings were trained on. We therefore keep random initialization of word embeddings for experiments on PHOENIX-Weather 2014T. Using this setting, we run a parameter search over the learning rate and warm-up steps, and we use initial learning rate 0.5 with 3,000 warm-up steps for the remaining experiments. Details of the parameter search are included in Appendix A.1.

Both GloVe and fastText English vectors have a reasonable overlap with the vocabulary of ASL glosses as well as the English targets (Table \ref{table:emb}). Therefore on ASLG-PC12 we load pre-trained embeddings on only the decoder, as well as on both the encoder and decoder.

\begin{center}
\resizebox{\linewidth}{!}{\begin{tabular}{lcccccccccccc}
\toprule
& \multicolumn{6}{c}{Dev Set} & \multicolumn{6}{c}{Test Set} \\
\cmidrule(l){2-7}\cmidrule(l){8-13}
 {Model} & {BLEU-1} & {BLEU-2}  & {BLEU-3} & {BLEU-4} & {ROUGE-L}  & {METEOR}& {BLEU-1} & {BLEU-2}  & {BLEU-3} & {BLEU-4} & {ROUGE-L}  & {METEOR}\\
 \midrule

Vanilla embedding & 90.15 & 84.92 & 80.27 & 75.94 & 94.72 & 94.58 & 90.49 & 85.64 & 81.31 & 77.33 & 94.75 & 95.16 \\ 
Tied dec & \bfseries 91.00 & \bfseries 86.26 & \bfseries  82.00 &\bfseries  78.02 & \bfseries 95.24 & \bfseries 95.12 & \bfseries 91.25 & \bfseries 86.76 & \bfseries 82.76 &\bfseries  79.02 & \bfseries  95.32 & \bfseries 95.75 \\ 
GloVe dec & 90.13 & 85.14 & 80.67 & 76.49 & 94.16 & 94.69 & 90.51 & 85.83 & 81.65 & 77.74 & 94.80 & 95.27 \\
GloVe enc-dec &  89.65 & 84.33 & 79.72 & 75.48 & 93.02 & 93.62 &90.01 & 85.15 & 80.88 & 76.95 & 93.00 & 94.14\\
fastText dec &90.64 & 85.63 & 81.14 & 76.94 & 94.75 & 95.02 & 91.20 & 86.62 & 82.53 & 78.72 & 94.73 & 95.57 \\
fastText enc-dec & 90.02 & 85.01 & 80.56 & 76.41 & 93.68 & 94.10  & 90.94 & 86.58 & 82.01 & 76.23 & 93.61 & 94.42 \\
fastText tied dec & 90.16 & 85.26 & 80.85 & 76.72 & 95.03 & 94.60 & 90.44 & 85.25 & 81.69 & 77.28 & 95.11 & 95.04 \\
\bottomrule
\end{tabular}} 
\label{table:aslemb}
 \captionof{table}{G2T performance comparison using different embedding schemes on ASLG-PC12.} 
\end{center}

Table 6 shows that fastText pre-trained embeddings for the decoder improves performance, and tied decoder embeddings with random initialization  gives  the  best performance. Weight tying is more suited on this dataset likely because it acts as regularization and combats overfitting, while the previous dataset is more complex and therefore less prone to overfitting. For the remaining experiments, we use tied decoder embeddings, initial learning rate 0.2 and 8,000 warm-up steps.

\subsubsection*{Beam width}

\begin{center}
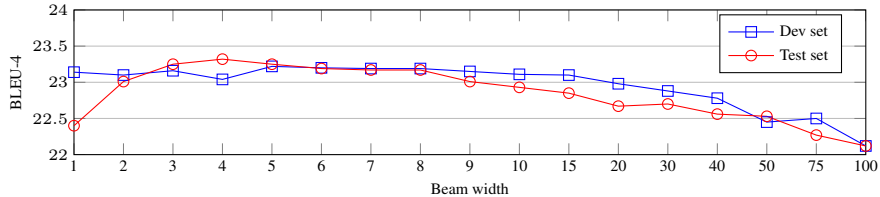

\tiny
\begin{tikzpicture}
\begin{axis}[
    xlabel={Beam width},
    ylabel={BLEU-4},
    width = 12cm,
    height = 3.5cm,
    ymajorgrids,
    xmin=0, xmax=160,
    ymin=22, ymax=24,
    xtick={0,10, 20, 30, 40, 50, 60, 70, 80, 90, 100, 110, 120, 130, 140, 150, 160},
    xticklabels={1, 2, 3, 4, 5, 6, 7, 8, 9, 10, 15, 20, 30, 40, 50, 75, 100},
    ytick={22,22.5,23,23.5,24},
    legend pos=north east,
]

\addplot[
    color=blue,
    mark=square,
    ]
    coordinates {
    (0, 23.14)(10,23.10)(20,23.16)(30,23.04)(40,23.22)(50,23.20)(60,23.19)(70,23.19)(80,23.15)(90,23.11)(100,23.10)(110,22.98)(120,22.88)(130,22.78)(140,22.45)(150,22.50)(160,22.12)
    };
\addplot[
    color=red,
    mark=o,
    ]
    coordinates {
    (0, 22.40)(10,23.01)(20,23.25)(30,23.32)(40,23.25)(50,23.19)(60,23.17)(70,23.17)(80,23.01)(90,22.93)(100,22.85)(110,22.67)(120,22.70)(130,22.56)(140,22.53)(150,22.27)(160,22.12)
    };
    \legend{Dev set, Test set}
    
\end{axis}
\end{tikzpicture}
\captionof{figure}{G2T decoding on RWTH-PHEONIX-WEATHER 2014T using different beam width.}
\label{fig:beam}
\end{center}

A naive method for decoding is greedy search, where the model simply chooses the word with the highest probability at each time step. However, this approach may become sub-optimal in the context of the entire sequence. Beam search addresses this by expanding all possible candidates at each time step and keeping a number of most likely sequences, or the beam width. Large beam widths do not always result in better performance and take more space in memory and decoding time. We search and find the optimal beam width value to be 4 on PHOENIX-Weather 2014T and 5 on ASLG-PC12.

\subsubsection*{Ensemble decoding}
Ensemble methods combine multiple models to improve performance. We propose ensemble decoding, where we combine the output of different models by averaging their prediction distributions. We chose 9 models from our experiments that gave the highest BLEU-4 during testing on PHOENIX-Weather 2014T. The number of models is chosen empirically, as using fewer models will lead to less ensembling but too many weaker models may lessen the quality of the ensemble model. These models are of the same architecture, but are initialized with different seeds and trained using different batch sizes and/or learning rates. These models give a BLEU-4 on testing between 22.92 and 23.41 individually.

\begin{center}
\resizebox{\textwidth}{!}{\begin{tabular}{lllllllllllll}
\toprule
& \multicolumn{6}{c}{Dev Set} & \multicolumn{6}{c}{Test Set} \\
\cmidrule(l){2-7}\cmidrule(l){8-13}
 {Model} & {BLEU-1} & {BLEU-2}  & {BLEU-3} & {BLEU-4} & {ROUGE-L}  & {METEOR}& {BLEU-1} & {BLEU-2}  & {BLEU-3} & {BLEU-4} & {ROUGE-L}  & {METEOR}\\
 \midrule

Raw data & 13.01 & 6.23 & 3.03 & 1.71 & 24.23 & 13.69 & 11.88 & 5.05 & 2.41 & 1.36 & 22.81 & 12.12 \\
RNN Seq2seq \cite{camgozslt} & 44.40 & 31.93 & 24.61 & 20.16 & 46.02 & \textendash & 44.13 & 31.47 & 23.89 & 19.26 & 45.45 & \textendash\\
Transformer \cite{camgoz2020sign} & \bfseries 50.69 & \bfseries 38.16& \bfseries 30.53 &\bfseries 25.35 & \textendash & \textendash & \bfseries 48.90 & 36.88 &29.45 &24.54 &\textendash & \textendash\\
Transformer & 49.05 & 36.20 & 28.53 & 23.52 & 47.36 & 46.09 & 47.69 & 35.52 & 28.17 & 23.32 & 46.58 & 44.85 \\
Transformer Ens. &  48.85 & 36.62 &  29.23 &  24.38 & \bfseries 49.01 & \bfseries 46.96 &  48.40 & \bfseries 36.90 & \bfseries 29.70 & \bfseries 24.90 & \bfseries 48.51 & \bfseries 46.24 \\
\bottomrule
\end{tabular}} 
 \captionof{table}{G2T on PHEONIX-WEATHER 2014T final results. \label{table:ensemble}
} 
\end{center}

Table \ref{table:ensemble} gives a performance comparison on PHOENIX-Weather 2014T of the recurrent seq2seq model by \newcite{camgozslt}, Transformer trained concurrently by \newcite{camgoz2020sign}, our single model, and ensemble model. We also provide the scores on the gloss annotations to illustrate the difficulty of this task. 

Without any additional training, ensembling improves testing performance by over 1 BLEU-4. Also, we report an improvement of over 5 BLEU-4 on the state-of-the-art. A single Transformer also gives an improvement of over 4 BLEU-4 more than the state-of-the-art, which shows the advantage of Transformers for SLT, as shown also in \newcite{camgoz2020sign}.

\begin{center}
\resizebox{\textwidth}{!}{\begin{tabular}{lllllllllllll}
\toprule
& \multicolumn{6}{c}{Dev Set} & \multicolumn{6}{c}{Test Set} \\
\cmidrule(l){2-7}\cmidrule(l){8-13}
 {Model} & {BLEU-1} & {BLEU-2}  & {BLEU-3} & {BLEU-4} & {ROUGE-L}  & {METEOR}& {BLEU-1} & {BLEU-2}  & {BLEU-3} & {BLEU-4} & {ROUGE-L}  & {METEOR}\\
 \midrule

Raw data & 54.60 & 39.67 & 28.92 & 21.16 & 76.11 & 61.25 & 54.19 & 39.26 & 28.44 & 20.63 & 75.59 & 61.65 \\
Preprocessed data & 69.25 & 56.83 & 46.94 & 38.74 & 83.80 & 78.75 & 68.82 & 56.36 & 46.53 & 38.37 & 83.28 & 79.06 \\
Seq2seq \cite{aslseq} & \textendash & \textendash & \textendash & \textendash & \textendash & \textendash & 86.7 & 79.5 & 73.2 & 65.9 & \textendash & \textendash\\
Transformer & \bfseries 92.98 &  \bfseries 89.09 & 83.55 & \bfseries 85.63 & 82.41 & 95.93 & 92.98 &\bfseries 89.09 & 85.63 & 82.41 & 95.87 & \bfseries96.46  \\
Transformer Ens. & 92.67 & 88.72 & \bfseries 85.22 &  81.93 & \bfseries 96.18 & \bfseries 95.95 & \bfseries 92.88 & 89.22 & \bfseries85.95 & \bfseries 82.87 & \bfseries 96.22 & 96.60 \\
\bottomrule
\end{tabular}} 
 \captionof{table}{G2T on ASLG-PC12 final results} 
\label{table:aslens}
\end{center}

We also use 5 of the best models from our experiments on ASLG-PC12 in an ensemble. Individually, these models obtain between 81.72 and 82.41 BLEU-4 on testing. Table \ref{table:aslens} shows that the performance of our single Transformer surpasses the recurrent seq2seq model by \newcite{aslseq} by over 16 BLEU-4. The ensemble model reports an improvement of 0.46 BLEU-4 over the single model. There is relatively less increase from ensembling possibly because there is less variance across different models.

\subsection{German Sign2Gloss2Text (S2G2T)}
In S2G2T, both gloss recognition from videos and its translation to text are performed automatically. \newcite{camgozslt} claims the previous G2T setup to be an upper bound for translation performance, since it simulates having a perfect recognition system. However, this claim assumes that the ground truth gloss annotations give a full understanding of sign language, which ignores the information bottleneck in glosses. \newcite{camgoz2020sign} hypothesizes that it is therefore possible to surpass G2T performance without using GT glosses, which we confirm in this section. 

We perform experiments on the PHOENIX-Weather 2014T dataset as it contains parallel video, gloss and text data. On the other hand, the ASLG-PC12 corpus does not have sign language video information.

\begin{center}
\resizebox{\textwidth}{!}{\begin{tabular}{lcccccccccccc}
\toprule
& \multicolumn{6}{c}{Dev Set} & \multicolumn{6}{c}{Test Set} \\
\cmidrule(l){2-7}\cmidrule(l){8-13}
 {Model} & {BLEU-1} & {BLEU-2}  & {BLEU-3} & {BLEU-4} & {ROUGE-L}  & {METEOR}& {BLEU-1} & {BLEU-2}  & {BLEU-3} & {BLEU-4} & {ROUGE-L}  & {METEOR}\\
 \midrule
 G2T \cite{camgozslt} & 44.40 & 31.93 & 24.61 & 20.16 & 46.02 & \textendash & 44.13 & 31.47 & 23.89 & 19.26 & 45.45 & \textendash\\
 S2G $\rightarrow $ G2T \cite{camgozslt} &  41.08 & 29.10 & 22.16 & 17.86 & 43.76 & \textendash &  41.54 & 29.52 & 22.24 & 17.79 & 43.45 & \textendash \\
  S2G2T  \cite{camgozslt}  & 42.88 & 30.30 & 23.03 & 18.40 & 44.14 & \textendash  & 43.29 & 30.39 & 22.82 & 18.13 & 43.80 & \textendash \\
  Sign2(Gloss+Text)  \cite{camgoz2020sign} & 47.26 & 34.40 & 27.05& 22.38 & \textendash & \textendash & 46.61 & 33.73& 26.19 &21.32 & \textendash & \textendash \\
 \midrule 
 S2G $\rightarrow $ G2T & 46.75 & 34.99 & 27.79 & 23.06 & 47.29 & 45.23 & 47.49 & 35.89 & 28.62 & 23.77 & 47.32 & 45.54 \\
  \midrule
   Bahdanau & 45.89 & 32.24 & 24.93 & 20.52 & 44.46 & 43.48 & 47.53 & 33.82 & 26.07 & 21.54 & 45.50 & 44.87  \\
 Luong & 45.61 & 32.54 & 26.33 & 21.00 & 46.19 & 44.93 & 47.08 & 33.93 & 26.31 &21.75& 45.66 &  44.84  \\
 \midrule
 
Transformer &  48.27 & 35.20 & 27.47 & 22.47 & 46.31 & 44.95 & 48.73 & 36.53 & 29.03 & 24.00 & 46.77 & 45.78  \\
Transformer Ens. &  \bfseries50.31 & \bfseries 37.60 & \bfseries 29.81 & \bfseries 24.68 & \bfseries 48.70 &\bfseries 47.45 & \bfseries 50.63 & \bfseries 38.36 & \bfseries 30.58 & \bfseries 25.40 & \bfseries 48.78 & \bfseries 47.60 \\
\bottomrule
\end{tabular}}
\label{table:stmc}
 \captionof{table}{SLT performance using STMC for CSLR. The first set of rows correspond to the current state-of-the-arts included for comparison. } 
\end{center}

\subsubsection*{ S2G $\rightarrow $G2T}

To begin, we use the best performing model for German G2T to translate glosses predicted by a trained STMC network. In Table 9 we can see that despite  no additional training for translation, this model already obtains a relatively high score that beats the current state-of-the-art by over 5 BLEU-4. 

\subsubsection*{Recurrent seq2seq networks}
For comparison, we also train and evaluate STMC used with recurrent seq2seq networks for translation. The translation models are composed of four stacked layers of Gated Recurrent Units (GRU) \cite{gru}, with either Luong \cite{luong} or Bahdanau \cite{bahdanau} attention.

In Table 9, recurrent seq2seq models obtain slightly better performance with Luong attention. Surprisingly, these models outperform previous models of similar architecture that translate GT glosses.

\subsubsection*{Transformer}

For the STMC-Transformer, we train Transformer models with the same architecture as in G2T. Parameter search yields an initial learning rate 1 with 3,000 warm-up steps and beam size 4. We empirically find using the 8 best models in ensemble decoding to be optimal. These models individually obtain between 23.51 and 24.00 BLEU-4.

Again, we observe that STMC-Transformer outperforms the previous system with ground truth glosses and Transformer. While STMC performs imperfect CSLR, its gloss predictions may be more useful than ground-truth annotations during SLT and are more readily analyzed by the Transformer. Again, the ground truth glosses represent merely a simplified intermediate representation of the actual sign language, so it is not entirely unexpected that translating ground truth glosses does not give the best performance.

STMC-Transformer also outperforms Transformers that translate GT glosses. While STMC performs imperfect CSLR, its gloss predictions may be better processed by the Transformer. Glosses are merely a simplified intermediate representation of the actual sign language so they may not be optimal. This result also reveals, training the recognition model to output more accurate glosses will not necessarily improve translation.

Both our STMC-Transformer and STMC-RNN also outperform \newcite{camgoz2020sign}'s model. Their best model jointly train Transformers for recognition and translation, however it obtains 24.49 WER on recognition whereas STMC obtains a better WER of 21.0, which suggests their model may be weaker in processing the videos. 

Moreover, Transformers outperform recurrent networks in this setup as well and STMC-Transformer improves the state-of-the-art for video-to-text translation by 7 BLEU-4.

\section{Qualitative comparison}

Example outputs of the G2T and S2G2T models (Table 10) show that the translations are of generally good quality, even with low BLEU scores. Most translations may have slight differences in word choice that do not change the overall meaning of the sentence or make grammatical errors, which suggests BLEU is not a good representative of human useful features for SLT. As for the comparison between the G2T and S2G2T networks, there does not seem to be a clear pattern between cases where S2G2T outperforms G2T and vice versa. One thing to note, though, is that the PHOENIX-Weather 2014T is restricted to the weather forecast domain, and a SLT dataset with a wider domain would be required to fully assess the performance of our model in more general real-life settings.

We also provide sample G2T outputs on the ASLG-PC12 corpus in Appendix A.2. 

\section{Conclusions and Future Work}
In this paper, we proposed Transformers for SLT, notably the STMC-Transformer. Our experiments demonstrate how Transformers obtain better SLT performance than previous RNN-based networks. We also achieve new state-of-the-art results on different translation tasks on the PHOENIX-Weather 2014T and ASLG-PC12 datasets.

A key finding is we obtain better performance by using a STMC network for tokenization instead of translating GT glosses. This calls into question current methods that use glosses as an intermediate representation, since reference glosses themselves are suboptimal. 

End-to-end training without gloss supervision is one promising step, though \newcite{camgoz2020sign}'s end-to-end model does not yet surpass their joint training model. As future work, we suggest continuing work on end-to-end training of the recognition and translation models, so the recognition model learns an intermediate representation that optimizes translation, or using a different sign language annotation scheme that has less information loss.

\begin{center}
\resizebox{.92\textwidth}{!}{\begin{tabular}{ll|c}
\toprule
 &  & BLEU-4 \\
 
 \midrule
    REF: & ähnliches wetter auch am donnerstag . & \\
    &\footnotesize (similar weather on thursday .) & \\
 G2T: & GLEICH WETTER AUCH DONNERSTAG  & \\
 & \footnotesize (SAME WEATHER ON THURSDAY) & \\
 & ähnliches wetter auch am donnerstag . & 100.00 \\
  & \footnotesize (similar weather on thursday .)& \\ 
 S2G2T: & GLEICH WETTER DONNERSTAG & \\
  & \footnotesize (SAME WEATHER THURSDAY) & \\
 & ähnliches wetter dann auch am donnerstag . & 48.89 \\
 & \footnotesize (similar weather then on thursday .)& \\ 
 
\midrule
  
       REF: & der wind weht meist schwach aus unterschiedlichen richtungen . & \\
        &\footnotesize (the wind usually blows weakly from different directions .) \\
 G2T: &  WIND SCHWACH UNTERSCHIED KOMMEN & \\
 & \footnotesize (WIND WEAK DIFFERENCE COME) & \\
 & der wind weht meist nur schwach aus unterschiedlichen richtungen . & 65.80 \\
 
  & \footnotesize (the wind usually blows only weakly from different directions .)& \\ 
  
 S2G2T: & WIND SCHWACH UNTERSCHIED & \\
  & \footnotesize (WIND WEAK DIFFERENCE) & \\
 &  der wind weht schwach aus unterschiedlichen richtungen . & 61.02\\
  
  & \footnotesize (the wind is blowing weakly from different directions .)& \\  
 
 \midrule
  
  REF: & sonnig geht es auch ins wochenende samstag ein herrlicher tag mit temperaturen bis siebzehn grad hier im westen . & \\
  &\footnotesize (the weekend is also sunny and saturday is a wonderful day with temperatures up to seventeen degrees here in the west .) \\
 G2T: & WOCHENENDE SONNE SAMSTAG SCHOEN TEMPERATUR BIS SIEBZEHN GRAD REGION &\\
   & \footnotesize (WEEKEND SUN SATURDAY NICE TEMPERATURE UNTIL SEVENTEEN DEGREE REGION)  & \\
   & und am wochenende da scheint die sonne bei temperaturen bis siebzehn grad . & 13.49 \\
   & \footnotesize (and on the weekend the sun shines at temperatures up to seventeen degrees .)& \\ 
 S2G2T: & WOCHENENDE SONNE SAMSTAG TEMPERATUR BIS SIEBZEHN GRAD REGION & \\
    & \footnotesize (WEEKEND SUN SATURDAY TEMPERATURE UNTIL SEVENTEEN DEGREE REGION) & \\
   & am wochenende scheint die sonne bei temperaturen bis siebzehn grad . & 12.55 \\
  &  \footnotesize (on the weekend sun shines at temperatures up to seventeen degrees .)& \\  
  
   \midrule
  
  REF: & es gelten entsprechende warnungen des deutschen wetterdienstes . & \\
    &\footnotesize (appropriate warnings from the german weather service apply .) \\
 G2T: & IX SCHON WARNUNG DEUTSCH WETTER DIENST STURM KOENNEN &\\
    & \footnotesize (IX ALREADY WARNING GERMAN WEATHER SERVICE STORM CAN) & \\
& es bestehen entsprechende unwetterwarnungen des deutschen wetterdienstes .  & 38.26 \\    
   &  \footnotesize (severe weather warnings from the german weather service exist .)& \\ 
 S2G2T: & DANN IX SCHON WARNUNG DEUTSCH WETTER STURM KOENNEN &\\
 & es gelten entsprechende warnungen des deutschen wetterdienstes . & 100.00 \\
     & \footnotesize (THEN IX ALREADY WARNING GERMAN WEATHER STORM CAN) & \\
     & \footnotesize (appropriate warnings from the german weather service apply .)& \\ 
 
 \midrule
 
 REF: & richtung osten ist es meist sonnig . & \\
     &\footnotesize (it is mostly sunny towards the east .) \\
 G2T: & OST MEISTENS SONNE &\\
      & \footnotesize (MOST EAST SUN) & \\
& im osten bleibt es meist sonnig . & 43.47 \\      
     & \footnotesize (in the east it mostly stays sunny .)& \\ 
     
 S2G2T: & OST REGION MEISTENS SONNE  & \\
 & im osten ist es meist sonnig . & 80.91 \\
  & \footnotesize (MOST REGION EAST SUN) &\\
  & \footnotesize (in the east it is mostly sunny .)& \\  
   \midrule
  
 REF: & am tag elf grad im vogtland und einundzwanzig grad am oberrhein . & \\
 &\footnotesize (during the day eleven degrees in vogtland and twenty one degrees in upper rhine .) \\
 G2T: & AM-TAG ELF VOGEL LAND & \\
 
   & \footnotesize (IN-THE-DAY ELEVEN BIRD LAND) & \\
& elf grad am oberrhein . & 18.74 \\   
  & \footnotesize (eleven degrees in upper rhine .)& \\
 S2G2T: & ELF VOGEL ZWANZIG & \\
  & \footnotesize (ELEVEN BIRD TWENTY) &\\
  & am tag elf grad im vogtland und zwanzig grad im vogtland . & 54.91 \\
  & \footnotesize (during the day eleven degrees in vogtland and twenty degrees in vogtland .)& \\
 
  \bottomrule
\end{tabular}} 
\label{table:compare}
 \captionof{table}{Qualitative comparison of G2T and S2G2T on RWTH-PHEONIX-WEATHER 2014T. Glosses are capitalized. REF refers to the reference German translation.} 
\end{center}

\section*{Acknowledgements}
The Titan X Pascal used for this research was donated by the NVIDIA Corporation. The authors would also like to thank Jean-Baptiste Rémy for the helpful discussions and feedback with various aspects of this work, and Hao Zhou for sharing the details of her previous work.

\bibliographystyle{coling}
\bibliography{coling2020}
\clearpage

\appendix
\section{Appendices}

\subsection{Experiments on German G2T learning rate}
\label{sec:lr}

A learning rate that is too low results in a notably slower convergence, but setting the learning rate too high risks leading the model to diverge. To prevent the model from diverging, we apply the Noam learning rate schedule where the learning rate increases linearly during the first training steps, or the warmup stage, then decreases proportionally to the inverse square root of the step number. The number of warmup steps is a parameter that has shown to influence Transformer performance \cite{transformertips} therefore we first run a parameter search over the number of warmup steps before finding the optimal initial learning rate.
\begin{center}
\tiny
\begin{tikzpicture}
\begin{axis}[
    xlabel={Warmup steps},
    ylabel={BLEU-4},
    width = 8.2cm,
    ymajorgrids,
    height = 4cm,
    xmin=0, xmax=70,
    ymin=21, ymax=23,
    xtick={0,10, 20, 30, 40, 50, 60, 70},
    xticklabels={1k, 2k, 3k, 4k, 5k, 6k, 7k, 8k},
    ytick={21,21.5,22,22.5,23},
    legend pos=south east,
]

\addplot[
    color=blue,
    mark=square,
    ]
    coordinates {
    (0, 21.72)(10, 22.49)(20, 22.24)(30,21.59)(40,21.40)(50,22.34)(60,22.30)(70,22.14)
    };
\addplot[
    color=red,
    mark=o,
    ]
    coordinates {
    (0, 21.81)(10, 22.22)(20, 22.41)(30,21.60)(40, 21.44)(50,21.92)(60,21.98)(70,22.20)
    };
    \legend{Dev set, Test set}
    
\end{axis}
\end{tikzpicture}
\captionof{figure}{G2T performance on RWTH-PHEONIX-WEATHER 2014T with different warmup steps. Initial learning rate is fixed to 0.2.}
\label{fig:warmup}

\tiny
\begin{tikzpicture}
\begin{axis}[
    xlabel={Learning rate},
    ylabel={BLEU-4},
    width = 8.2cm,
    height = 4cm,
    ymajorgrids,
    xmin=0, xmax=11,
    ymin=21, ymax=23,
    xtick={0,1, 2, 3, 4, 5, 6, 7, 8, 9, 10, 11},
    xticklabels={0.01, 0.05, 0.1, 0.2, 0.3, 0.4, 0.5, 0.6, 0.7, 0.8, 0.9, 1},
    ytick={21,21.5,22,22.5,23},
    legend pos=south east,
]

\addplot[
    color=blue,
    mark=square,
    ]
    coordinates {
    (0,21.54)(1,22.79)(2,22.83)(3,22.74)(4,22.33)(5,22.50)(6,22.83)(7,22.36)(8,22.43)(9,22.21)(10,22.01)(11,21.65)
    };
\addplot[
    color=red,
    mark=o,
    ]
    coordinates {
    (0,21.87)(1,22.37)(2,22.05)(3,22.41)(4,22.20)(5,22.17)(6,22.63)(7,22.50)(8,22.59)(9,22.48)(10,22.39)(11,22.19)
    };
    \legend{Dev set, Test set}
    
\end{axis}
\end{tikzpicture}
\captionof{figure}{G2T performance on RWTH-PHEONIX-WEATHER 2014T with various initial learning rate.}
\label{fig:lr}
\end{center}

 \subsection{Qualitative G2T Results on ASLG-PC12}
 Because quantitative metrics provide only a limited evaluation of translation performance, manual evaluation by viewing the translation outputs directly may give a better assessment of the quality of translations. 
Table \ref{table:aslqual} provides examples of SLT output on the ASLG-PC12 dataset. Here we can see how ASL glosses include prefixes that are not necessary to encapture the meaning of the phrase, which we have removed during data pre-processing before training. With a BLEU-4 testing score of 82.87, most predictions by our system are very close to the target English phrases and are able to convey the same meaning. We have also selected translation examples with lower BLEU-4 score and we can see that common errors include mistranslation of numbers and proper nouns. These are likely corner cases with infrequent examples during training.

\begin{center}
\resizebox{\textwidth}{!}{\begin{tabular}{rl|c}
\toprule
 & & BLEU-4 \\
 \midrule
ASL: & X-I BE DESC-PARTICULARLY DESC-GRATEFUL FOR EUROPEAN PARLIAMENT X-POSS DRIVE ROLE WHERE BALTIC SEA COOPERATION BE CONCERN & \\
GT: & i am particularly grateful for the european parliament's driving role where the baltic sea cooperation is concerned . & 100.00 \\
Pred: & i am particularly grateful for the european parliament's driving role where the baltic sea cooperation is concerned . & \\
\midrule
ASL: & DESC-REFORE , DESC-MUCH WORK NEED TO BE DO IN ORDER TO DESC-FURR SIMPLIFY RULE & \\
GT: & therefore , much work needs to be done in order to further simplify the rules . & 100.00 \\
Pred: & therefore , much work needs to be done in order to further simplify the rules . & \\
\midrule
ASL: & THIS PRESSURE BE DESC-PARTICULARLY DESC-GREAT ALONG UNION X-POSS DESC-SOURN AND DESC-EASTERN BORDER & \\
GT: & this pressure is particularly great along the union's southern and eastern borders . & 100.00 \\
Pred: & this pressure is particularly great along the union's southern and eastern borders . & \\
\midrule
ASL: & MORE WOMAN DIE FROM AGGRESSION DESC-DIRECT AGAINST X-Y THAN DIE FROM CANCER . & \\
GT: & more women die from the aggression directed against them than die from cancer . & 73.15 \\
Pred: & more women die from aggression directed against them than die from cancer . & \\
\midrule
ASL: & X-IT FUEL WAR IN CAMBODIUM IN 1990 AND X-IT BE ENEMY DEMOCRACY & \\
GT: & it fuelled the war in cambodia in the 1990s and it is the enemy of democracy . &  25.89 \\
Pred: & it fuel war in the cambodium in 1990 and it is an enemy of democracy . & \\
\midrule
ASL: & DESC-N CHIEF INVESTIGATOR X-HIMSELF BE TARGET AND HOUSE CARD COLLAPSE . & \\
GT: & then the chief investigator himself is targeted and the house of cards collapses . & 21.29\\
Pred: &then chief investigator himself is a target and a house card collapse . &  \\
\midrule
ASL: & U , X-WE TAKE DESC-DUE NOTE X-YOU OBSERVATION . AMENDMENT THANK X-YOU MR & \\
GT: & otherwise we have to vote on the corresponding part of amendment thank you mrs ţicău , we take due note of your observation . & 15.93  \\
Pred: & mr president , we took due note of your observation . & \\

\bottomrule
\end{tabular}} 
 \captionof{table}{Examples of ASL translation with varying BLEU-4 scores} 
\label{table:aslqual}
\end{center}

\end{document}